\documentclass[10pt]{iopart}
\usepackage{iopams}

\usepackage{graphicx}
\usepackage{subfigure}
\usepackage[bookmarks=false]{hyperref}
\hypersetup{colorlinks,allcolors=black}

\DeclareUnicodeCharacter{2009}{\,} 

\begin{document}

\title[Compliant AI Image-Based Detachment]{Regulation Compliant AI for Fusion: Real-Time Image Analysis-Based Control of Divertor Detachment in Tokamaks}

\author{Nathaniel Chen$^1$, Cheolsik Byun$^1$, Azarakash Jalalvand$^1$, Sangkyeun Kim$^2$, Andrew Rothstein$^1$, Filippo Scotti$^3$, Steve Allen$^3$, David Eldon$^4$, Keith Erickson$^2$, and Egemen Kolemen$^{1,2}$}

\address{$^1$ Princeton University, Princeton, New Jersey, USA 08540 \
$^2$ Princeton Plasma Physics Labratory, Princeton, New Jersey, USA 08540 \
$^3$ Lawrence Livermore National Laboratory, Livermore, California, USA \
$^4$ General Atomics, San Diego, California, USA}

\ead{\{ekolemen, nathaniel\}@princeton.edu}
\vspace{10pt}
\begin{indented}
\item[]November 2024
\end{indented}

\begin{abstract}
While artificial intelligence (AI) has been promising for fusion control, its inherent black-box nature will make compliant implementation in regulatory environments a challenge. This study implements and validates a real-time AI enabled linear and interpretable control system for successful divertor detachment control with the DIII-D lower divertor camera. Using $D_2$ gas, we demonstrate feedback divertor detachment control with a mean absolute difference of 2\% from the target for both detachment and reattachment. This automatic training and linear processing framework can be extended to any image based diagnostic for regulatory compliant controller necessary for future fusion reactors.
\end{abstract}

\vspace{2pc}

\ioptwocol
\section{Introduction}

Fusion plasmas hold great promise as an abundant source of clean energy. However, their inherently nonlinear behavior poses significant challenges to classical control methods. Recent advances in artificial intelligence (AI) have demonstrated that complex plasma phenomena can be controlled by uncovering hidden patterns in high-dimensional data spaces, which are otherwise difficult to derive \cite{rothstein_initial_2024, seo_avoiding_2024, kim_disruption_2024}. Although these AI-based approaches open exciting avenues for advanced control, the black-box nature of many algorithms raises safety concerns for deployment in next-generation fusion power plants (FPP). This motivates the development of safe, interpretable AI control methods that are crucial for the practical realization of fusion energy.

A critical area for this is the active control of divertor detachment, essential for protecting plasma-facing components (PFCs) from excessive kinetic plasma energy. Diagnostics used for reliable detachment measurement often rely on sparse sampling, as these measurements are physically interpretable and practical for implementation. However, the sparsity of these measurements physically constrains their control capabilities, limited by the location of the plasma. In devices such as DIII-D, KSTAR, EAST, JET, ASDEX-U, and others, Langmuir probes measure the current saturation-based fractional divertor detachment parameter, $A_{frac} = J_{sat}/J_{roll}$, at discrete points along the divertor surface \cite{wang_integration_2021, eldon_enhancement_2022, guillemaut_real-time_2017}. High burst-rate lasers used for Thomson scattering can measure electron temperature ($T_e$) at kilohertz sampling rates along a single chord \cite{funaba_electron_2022, ravensbergen_real-time_2021}. Line-integrated bolometers measure heat flux along individual chords passing through the bolometer, enabling direct heat computation \cite{peterson_signal_2018}.

\begin{figure}
    \includegraphics[width=\columnwidth]{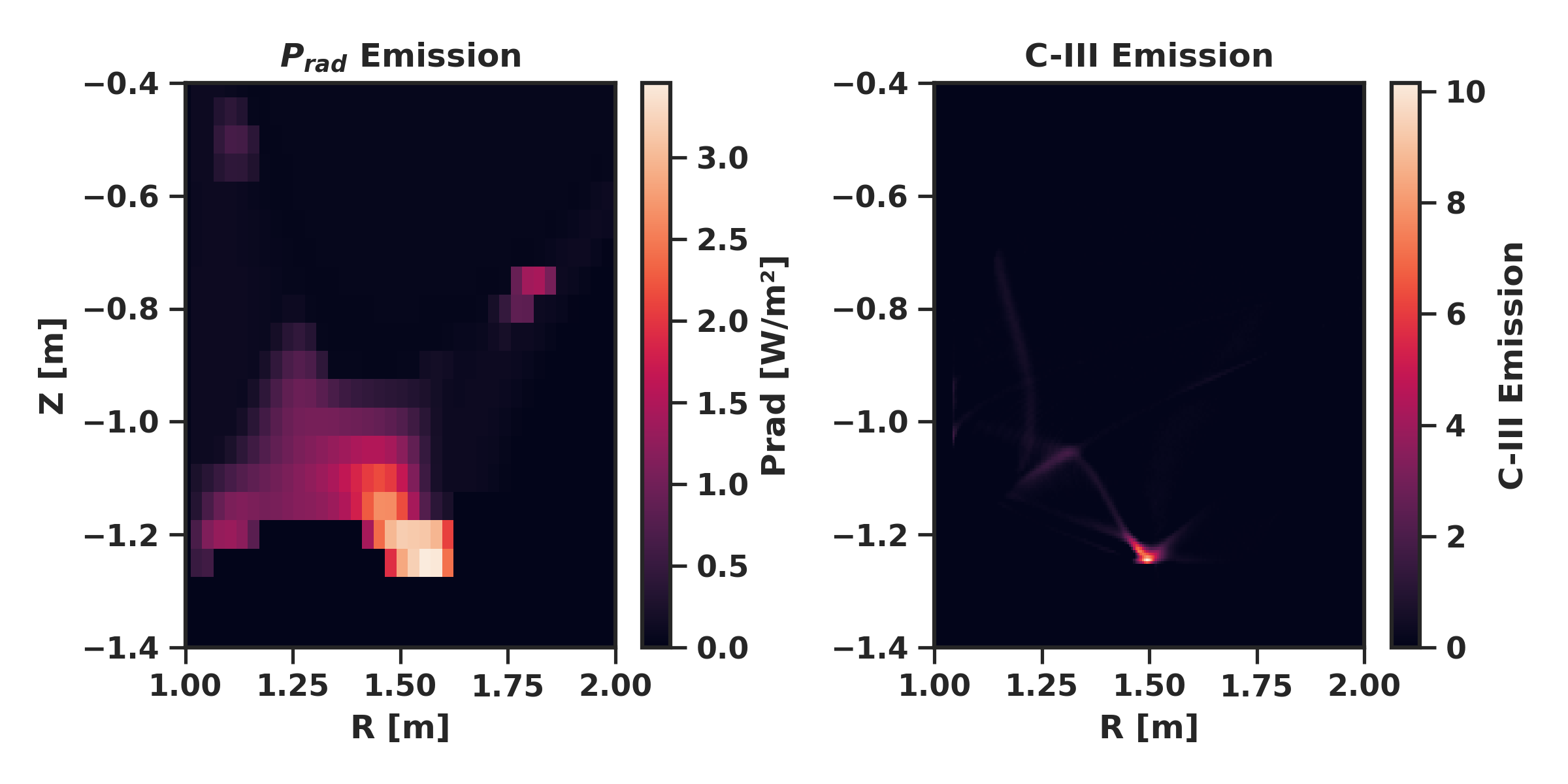}
    \caption{DIII-D 201081 at 2700ms shows a comparison between the tomographically inverted C-III visible distribution (blue) and $P_{rad}$ distribution (red) at DIII-D. $P_{rad}$ is an integrated measurement near divertor. In comparison, $DZ$ from C-III is positional, leveraging averaging algorithms.}
    \label{fig:pradciii}
\end{figure}

Diagnostics with high spatial resolution exist, such as the high-resolution imaging bolometers on devices like KSTAR or the tangential camera system (TangTV) at DIII-D \cite{leonard_2d_1995, fenstermacher_tangentially_1997}. A comparison between the the resolution of the camera and bolometry system at DIII-D can be seen in Figure \ref{fig:pradciii}. Although these have been used for control, they typically operate under limited scenarios and approximations. For example, experiments on the TCV tokamak demonstrated active plasma detachment control using the MANTIS diagnostic with an approximated parametric view of Carbon-III (C-III) emission gradients \cite{ravensbergen_real-time_2021}. However, in devices like DIII-D, such approximations become complicated due to the tokamak's short leg length, altering the emissivity profile near the strike point. Although calculating physical location points for detachment control is feasible, real-time high-resolution tomographic reconstruction remains challenging; for instance, accurate inversion on DIII-D may require iterative schemes that is hard to achieve in real time \cite{carr_description_2018}.

To address this, machine learning (ML) offers a promising path forward, enabling models that empirically map labeled detachment values directly to camera viewpoints from diverse setups, thus bypassing uncertainties associated with direct image estimation methods and manual nonlinear modeling. Convolutional neural networks (CNNs) have successfully regressed emission fronts from DIII-D data \cite{boyer_neural_2024, victor_identifying_2024}, and AI-based detachment control has been implemented on devices like KSTAR \cite{zhu_latent_2025}. Nonetheless, the black-box or grey-box nature of many ML models remains a challenge for integration into highly regulated systems.

In response, this study proposes a visually interpretable linear control model trained using ML techniques, aiming to facilitate safer system integration while maintaining robust performance. Utilizing such white-box, interpretable algorithms paves the way for practical implementation and operation in ITER and fusion pilot plants (FPP) \cite{dennis_artificial_2023}. An example of this can be with interpretable ML for disruption prediction using random forests \cite{rea_real-time_2019}. Our analysis shows that while ML is indeed valuable for designing control systems, deep neural networks are not always necessary for real-time operation. ML-trained linear systems can effectively meet many fusion control objectives \cite{goodfellow_deep_2016}.

The rest of the paper is structured as follows:

\begin{enumerate}
    \item Experimental Setup and Feedback Control Scheme: A description of the diagnostics, controller, and machine learning setup.
    \item Results of Deuterium Seeding Control Experiments: Experiments with L-Mode control, experiments with H-mode control, post-shot analysis, and a comparison between modifications of the control variable.
\end{enumerate}

\section{Experimental Setup and Feedback Control Scheme}

\subsection{Setup and Diagnostics}

The experiments were conducted at the DIII-D National Fusion Facility, focusing on lower single-null divertor configurations in standard L-mode and H-mode scenarios with only Neutral Beam Injection (NBI) as the heating source and Deuterium ($D_2$) gas puffing as the actuation. Shots are referenced off of L-Mode density ramp shot 199166 with reference feed forward gas A waveforms as the base case from 199349. Table \ref{tab:t1} shows the range of parameters for each shot.

\begin{table*}[t]
\centering
\begin{tabular}{rrrrrl}
\hline
\multicolumn{1}{l}{Shot}       & \multicolumn{1}{l}{IP} & \multicolumn{1}{l}{BTOR} & \multicolumn{1}{l}{PBEAM (MW)} & \multicolumn{1}{l}{KAPPA} & GAS    \\ \hline
189057                            & 0.93                   & -2.09                    & 8.82                           & 1.69                      & D2, N2 \\ \hline
189061                            & 0.92                   & -2.1                     & 8.75                           & 1.72                      & N2     \\ \hline
189062                            & 0.93                   & -2.09                    & 8.92                           & 1.7                       & N2     \\ \hline
189081                            & 1.54                   & -2.09                    & 10.97                          & 1.71                      & N2     \\ \hline
189088                            & 1.53                   & -2.1                     & 11.08                          & 1.7                       & N2     \\ \hline
189090                            & 1.54                   & -2.1                     & 11.14                          & 1.7                       & N2     \\ \hline
189093                            & 0.93                   & -2.09                    & 11.05                          & 1.84                      & N2     \\ \hline
189094                            & 0.93                   & -2.09                    & 11.16                          & 1.83                      & N2     \\ \hline
189097                            & 0.94                   & -2.1                     & 11.03                          & 1.73                      & Ne     \\ \hline
189100                            & 0.91                   & -2.07                    & 11.07                          & 1.74                      & Ne     \\ \hline
189101                            & 0.93                   & -2.08                    & 11.1                           & 1.73                      & Ne     \\ \hline
189448                            & 1.29                   & -2.07                    & 2.25                           & 1.44                      & N2     \\ \hline
189451                            & 1.29                   & -2.06                    & 2.26                           & 1.62                      & N2     \\ \hline
190109                            & 1.29                   & -2.09                    & 1.86                           & 1.62                      & N2     \\ \hline
190110                            & 1.29                   & -2.09                    & 1.8                            & 1.62                      & N2     \\ \hline
190113                            & 1.32                   & -2.08                    & 1.8                            & 1.71                      & N2 \\ \hline
190114                            & 1.34                   & -2.09                    & 1.83                           & 1.63                      & N2 \\ \hline
190115                            & 1.3                    & -2.09                    & 1.83                           & 1.69                      & N2 \\ \hline
190116                            & 1.3                    & -2.08                    & 1.83                           & 1.62                      & N2 \\ \hline
199166                            & 0.94                   & -2.1                     & 1.78                           & 1.63                      & D2, N2 \\ \hline
199171                            & 0.91                   & -2.09                    & 1.96                           & 1.62                      & D2, N2 \\ \hline
199172                            & 0.91                   & -2.09                    & 1.91                           & 1.63                      & D2, N2 \\ \hline
199351-199354 & 0.91                   & -2.09                    & 1.71-1.75  & 1.63                      & D2, N2 \\ \hline
\end{tabular}
\caption{List of training shot and parameters. This includes shots with L-Mode, H-Mode, and MARFES.}
\label{tab:t1}
\end{table*}

The  diagnostic used for control was the TangTV system, which captures images of C-III emission at a wavelength of 465 nm using a bandpass filter.

\begin{figure*}
\centering
\includegraphics[width=0.75\textwidth]{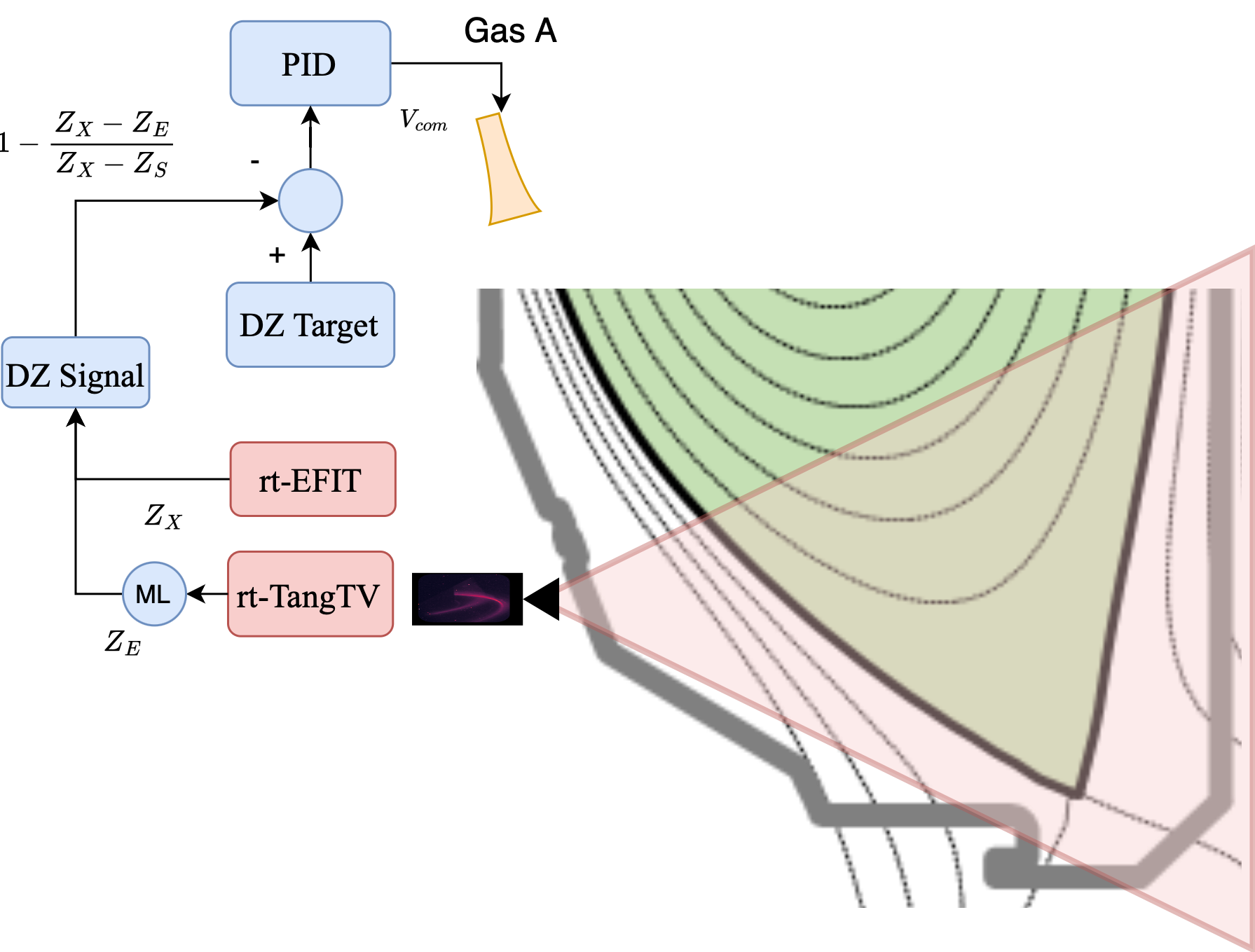}
\caption{Active control feedback loop at DIII-D for rt-TangTV control of $DZ$ for detachment. A ML model extracts $Z_E$ from the lower divertor C-III image, then $DZ$ signal is calculated by finding the ratio between this value and the value of the $X$ point and strike point. Strike point is manually set to the height to the lower divertor.}
\label{fig:setup}
\end{figure*}

\subsection{Diagnostics and Data Acquisition}
Analog signals were acquired using the Thermo Fisher CID 8725D1M Camera. Signals were digitized via a PIXCI A310 Frame Grabber before being sent to the Plasma Control System (PCS) \cite{margo_current_2020}. The camera is mounted outside the vessel tangentially to the lower divertor port, and records at a resolution of 720 x 480. Edge localized modes (ELMs) are integrated over the 60Hz frame rate of the camera (full frames, deinterlaced to 30~Hz), giving a consistent signal.

The camera is mounted outside the lower divertor port, providing a tangential view of the divertor region (see Figure \ref{fig:setup}). The images were converted to digital signals using a frame grabber configured to capture full frames at 30 Hz. X-point locations were taken from real time Equilibrium Fitting (rt-EFIT) \cite{lao_reconstruction_1985}.

\subsection{Control Input Metric}

\begin{figure}
    \includegraphics[width=\columnwidth]{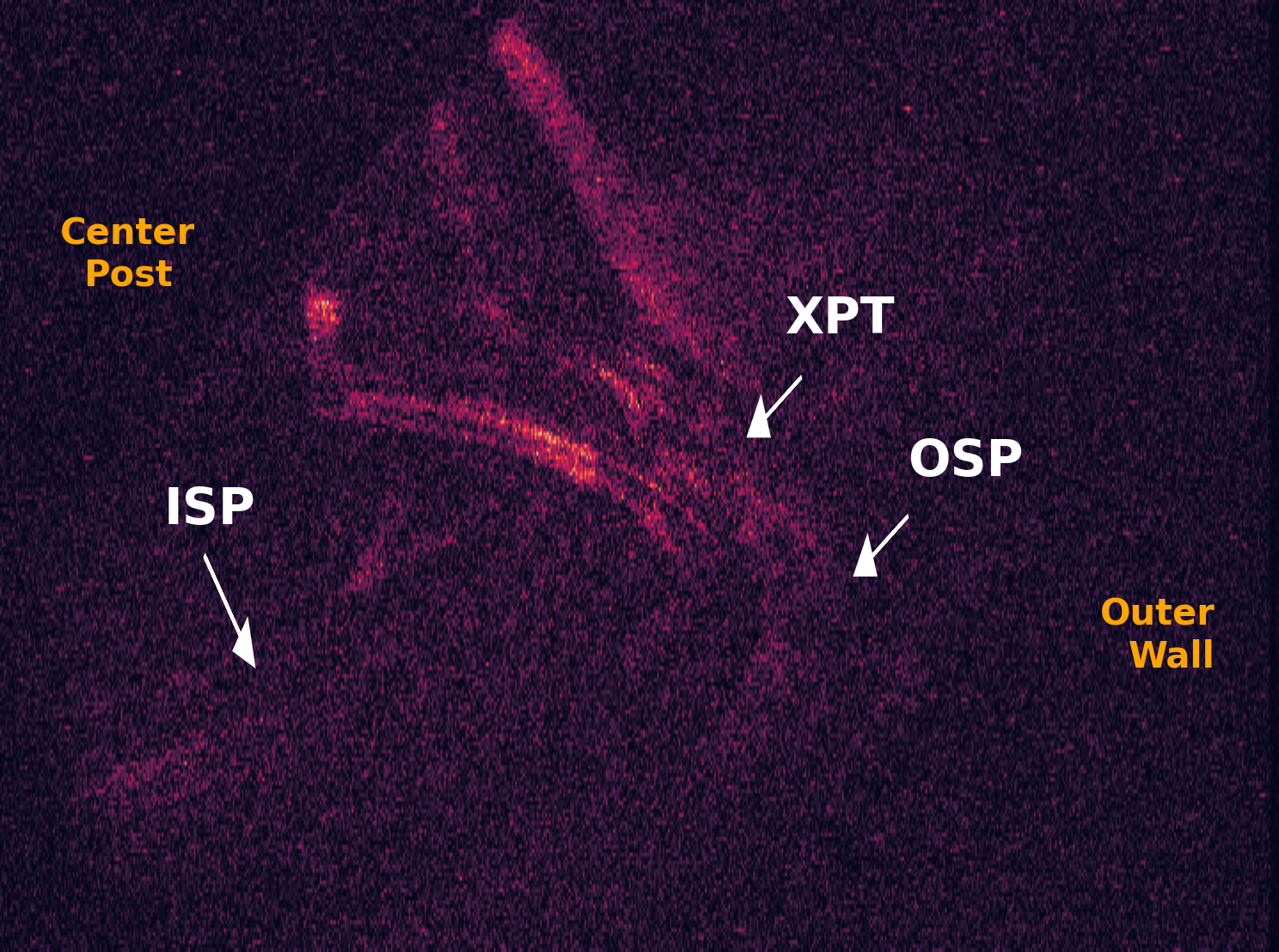}
    \caption{Linear Activation map of MDL-Hist shows the impact of both the ring and leg on $Z_E$ prediction. $Z_E$ is calculated by taking the dot product of this with a camera image.}
    \label{fig:activation}
\end{figure}

The control aim was to create a single-input-single-output (SISO) model that could track continuous changes. In order to create an interpretable input to the model given an image, we could take advantage of the fact that as C-III follows a certain $T_e$, cooling of the divertor region leads to the changes in location fo the C-III emission front along the parametric viewpoint, termed $Z_E$. The images were used in a linear regression model for faster processing on the PCS CPU. The model used in this case is \ref{fig:activation}, which requires only taking the dot product with the real time image to estimate the label. The final control input used for feedback control in the PCS $DZ$, the normalized relative to the vertical distance from the X-point to the strike point. The equation for $DZ$ is

\begin{equation}
    DZ = 1 - \frac{Z_X - Z_E}{Z_X - Z_S}
\end{equation}

where $Z_X$ is the vertical position of the X-point, and $Z_S$ is the vertical position of the strike point. Interpreting $DZ$:

$DZ = 0$: $Z_E$ is at the strike point.

$DZ = 1$: $Z_E$ is at the X-point.

$DZ > 1$: $Z_E$ is above the X-point, indicating the radiation front is above the strike point.

\subsection{Determining $Z_E$ Labels}

\begin{figure*}
\subfigure[]{\includegraphics[width=0.49\textwidth]{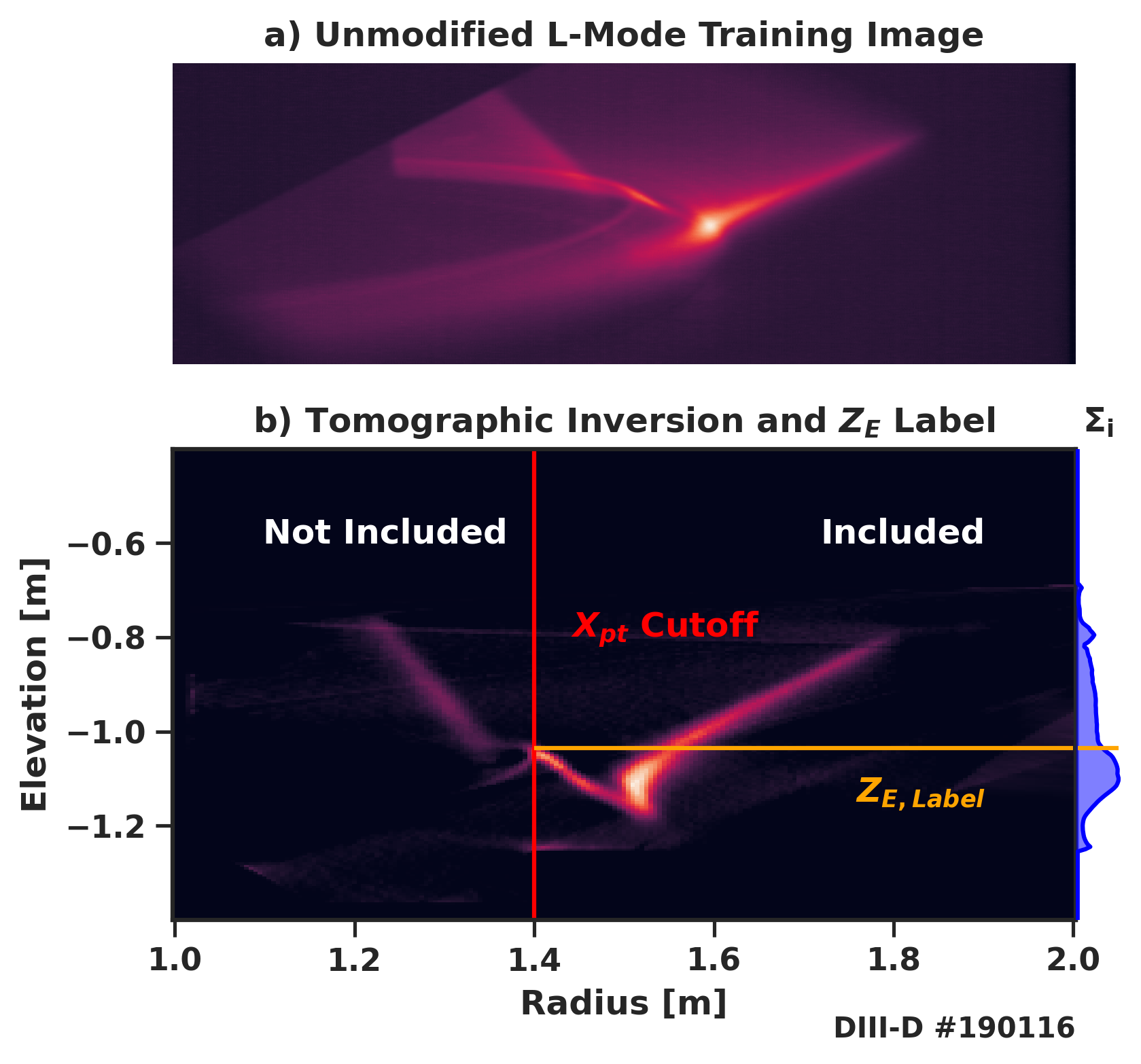}} 
\subfigure[]{\includegraphics[width=0.49\textwidth]{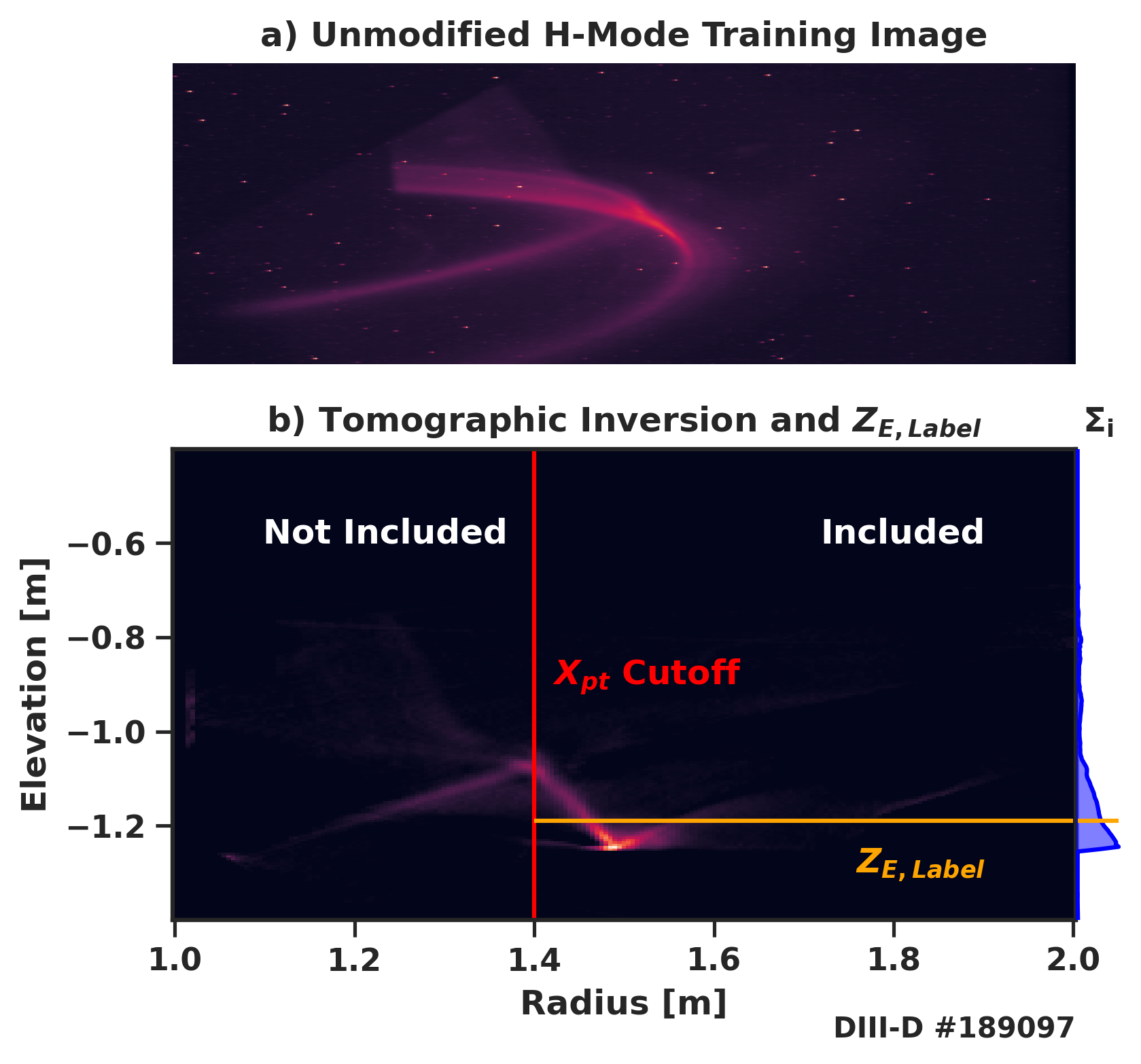}}
\caption{Average intensity labeling scheme first sums all pixels right (outboard) of the X-point inversion and then finds the root mean square elevation (a) In L-Mode (b) In H-Mode}
\label{fig:camera}
\end{figure*}

Labeled $Z_E$, dubbed $Z_{E,Label}$, were created from 3,500 tomographically inverted images \cite{meyer_tomographic_2018}. The mean elevation value of the emission intensity of all pixels to the right (outboard) of the X-point was calculated to represent the emission height of the OSP. Our training dataset included H-mode scans, L-mode scans, and phenomena such as Multifaceted Asymmetric Radiation from the Edge (MARFEs) as well as forward and reverse $B_T$. MARFEs are a type of thermal instability that occur when the highly collisional area moves near the X-Point. Both $B_T$ directions are important to include in training because the intensity profile in the outer leg varies significantly with $B_T$ direction. This method provides a robust labeling of detachment levels applicable across both H-mode and L-mode configurations. We define $Z_{E,Label}$ as

\begin{equation}
    Z_{E,Label} = \sqrt{\frac{(\sum_{i=0}^{n}((i \sum_{j=j_x}^{m} I_{ij}))^2}{\sum_{i=0}^{n}( \sum_{j=j_x}^{m} I_{ij})^2}}
\end{equation}

where in pixel space where $I_{ij}$ is the pixel intensity at position $(i,j)$, $n$ is the number of rows, $m$ is the number of columns, and $j_x$ is the column index corresponding to the X-point. Finally, the average pixel coordinate is transformed back into meters, results shown in Figure \ref{fig:camera}.

This approach avoids issues with unclear emission point labeling or leg length determination and requires no additional preprocessing. It can also be noted that this algorithm is generalizable enough to be used for other types of spatial sensors that can be used to quantify detachment. It is also important to note that $Z_{E,Label}$ is not ready to be used in real-time since it comes from the tomographic inverted images which have a high computational cost to generate. Therefore the ML will be used to directly translate the camera frame into an inferred $Z_E$. However, it turns out that additional processing is required to resolve inconsistencies in how the ML model infers $Z_E$ when used on the real-time camera.

\subsection{ML-inferred $Z_E$ Processing for Control}

\begin{figure}
    \includegraphics[width=\columnwidth]{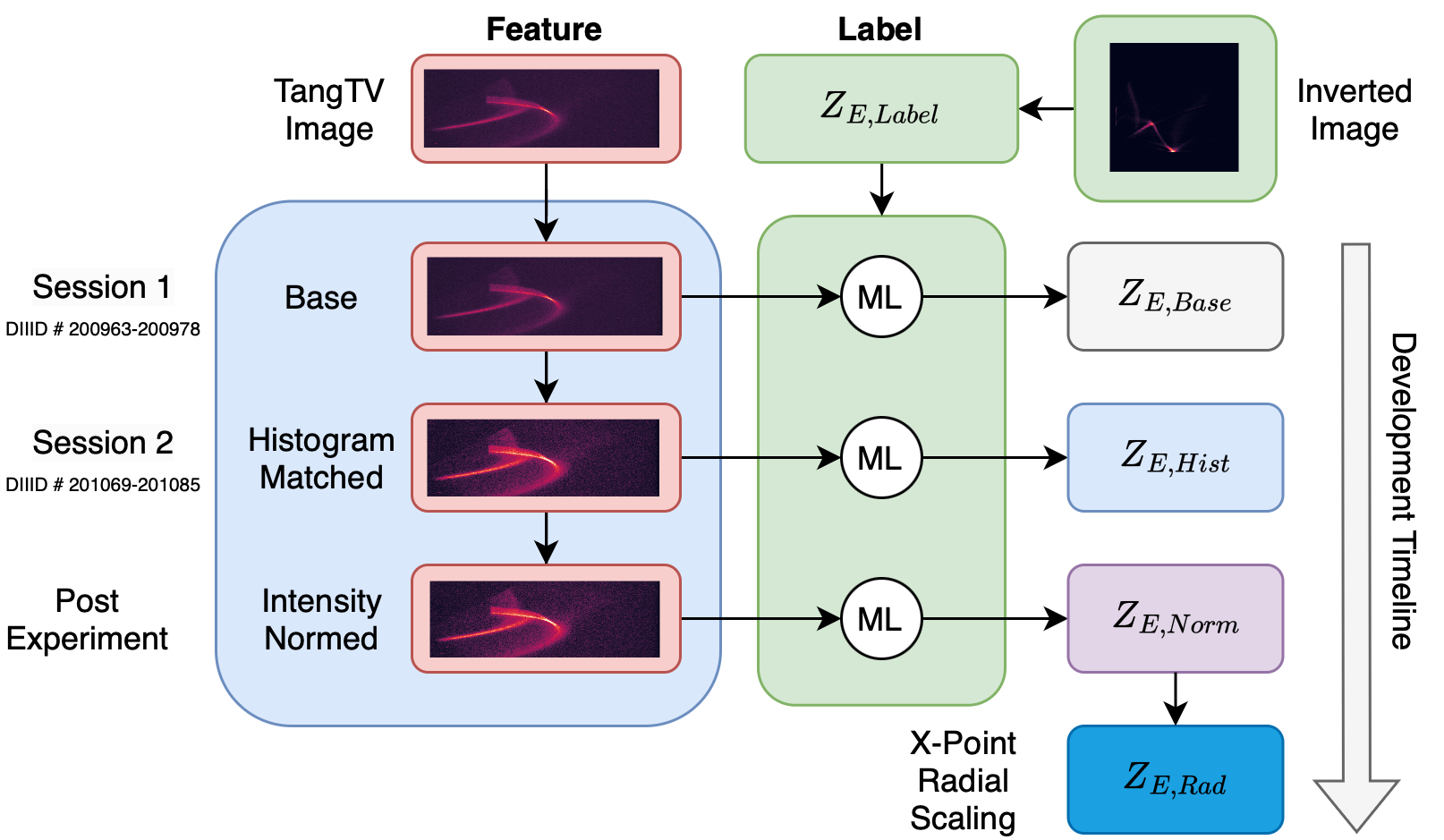}
    \caption{The TangTV images used for training was determined to have a different intensity profile from the rt-TangTV profile during the first half of experiments (DIII 200963-200978) using $Z_{E,Base}$. Subsequent analysis of these shots showed that matching training images to the histogram of the rt-TangTV for DIII 201069-201085 gave more accurate results, shown by $Z_{E,Hist}$. After the experiment, $Z_{E,Norm}$ was created to remove dependence of $Z_E$ on the global brightness of the plasma. Finally $Z_{E,Rad}$ subtracts a scaled quantity of the radial component of the X point, found in Equation \ref{eq:rdz}.}
    \label{fig:preprocessing}
\end{figure}

$Z_E$ was developed throughout the experiment to address differences in real-time compared to simulated predictions. Figure \ref{fig:preprocessing} shows the the different $Z_E$ metrics. While For the first set of experiments, shots 200963-200978, unprocessed historical TangTV values were used for to train a model that would infer $Z_{E,Base}$.

However, rt-TangTV have a different intensity profile from the historical TangTV data due to signal variations from background subtraction. To account for this, additional preprocessing steps were taken.

where $\alpha$ is the noise amplitude, and $N$ is a normal distribution.

The histograms of the training images were matched to the real-time camera profiles to ensure consistency in brightness and contrast \cite{morovic_fast_2002}. These two combine to form a model that would infer $Z_{E,Hist}$, shown in Figure \ref{fig:activation}. This value was subsequently used in shots 201069-201085.

In addition to intensity preprocessing, speckle noise was added to simulate the effect of radiation on sensor:
\begin{equation}
    I_{noise} = I + \alpha \times N
\end{equation}

\subsection{System Identification}

System identification ensures that desirable control over the actuation is achieved given the input. One way to achieve this is through a linear Proportial-Integral-Derivative (PID) controller, which aims to minimize response delay and overshoot by accounting for the existing state of the system. Tuning is done by calculating initial PID gains from voltage step shots and then refining them with subsequent control shots.

In this experiment, system identification is performed by stepping the gas voltage command ($V_{com}$), fitting the $DZ$ response to a First Order Plus Dead Time (FOPDT). The FOPDT analysis and initial PID tuning calculation can be performed between shots using a dedicated utility in the OMFIT PCS module \cite{meneghini_integrated_2015, eldon_controlling_2017}.

\section{Results of Deuterium Seeding Control Experiments}

\subsection{Active Feedback Control in  L-Mode}

\begin{figure}
    \includegraphics[width=\columnwidth]{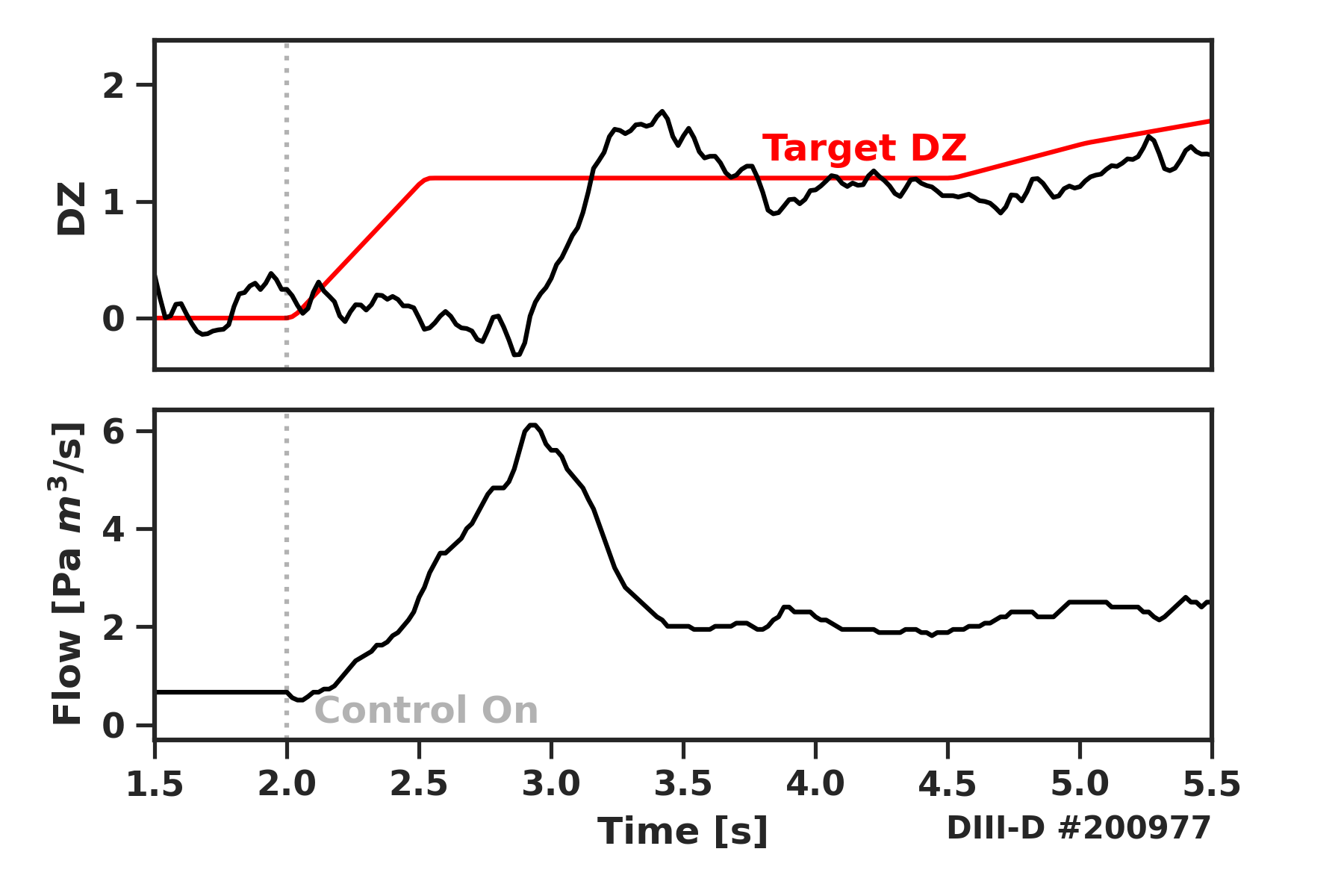}
    \caption{L-mode shot 200977 is controlled with $DZ_{Base}$, with increasing gas followed by an increasing measured $DZ$. The response delay is likely due to the gas penetration time. Notice that $DZ$ goes significantly above 1, which physically means that the most of the emission is above the X point. This error in measurement is later determined to be due to different intensity profiles from the training and rt-TangTV, which was subsequently adjusted for with $Z_{E,Hist}$. $DZ$ can be noticed becoming briefly negative from 2.0s to 3.0s. This is likely a limitation of the model used, as the $DZ$ increases in other models as shown in Figure \ref{fig:mdlcompare}}
    \label{fig:200977}
\end{figure}

Figure \ref{fig:200977} shows control over $DZ_{Base}$ signal in L-mode using $Z_{E,Base}$. Optimal PID gains that minimized oscillation while quickly increasing $DZ$ for this scenario were tuned such that the proportional gain $G_p = -1$, integral gain ratio $G_I/G_p = 5$, and derivative gain ratio $G_D/G_p = 250$. The low pass control filter was set to 40ms with manual adjustments based on past camera-based detachment studies at KSTAR. Nonlinear behavior was observed for control nearing the X-point. Previously, this phenomenon has been noted during marginal detachment. During shot 200997, it was observed that as the system approached the X-point (i.e., $DZ = 1$), the gas density fell off at time $t=2.9$s. Detachment was not the objective of this shot; we only aimed at controlling $DZ$ using rt-TangTV.

\subsection{Active Feedback Control in  H-Mode}

\begin{figure}
    \includegraphics[width=\columnwidth]{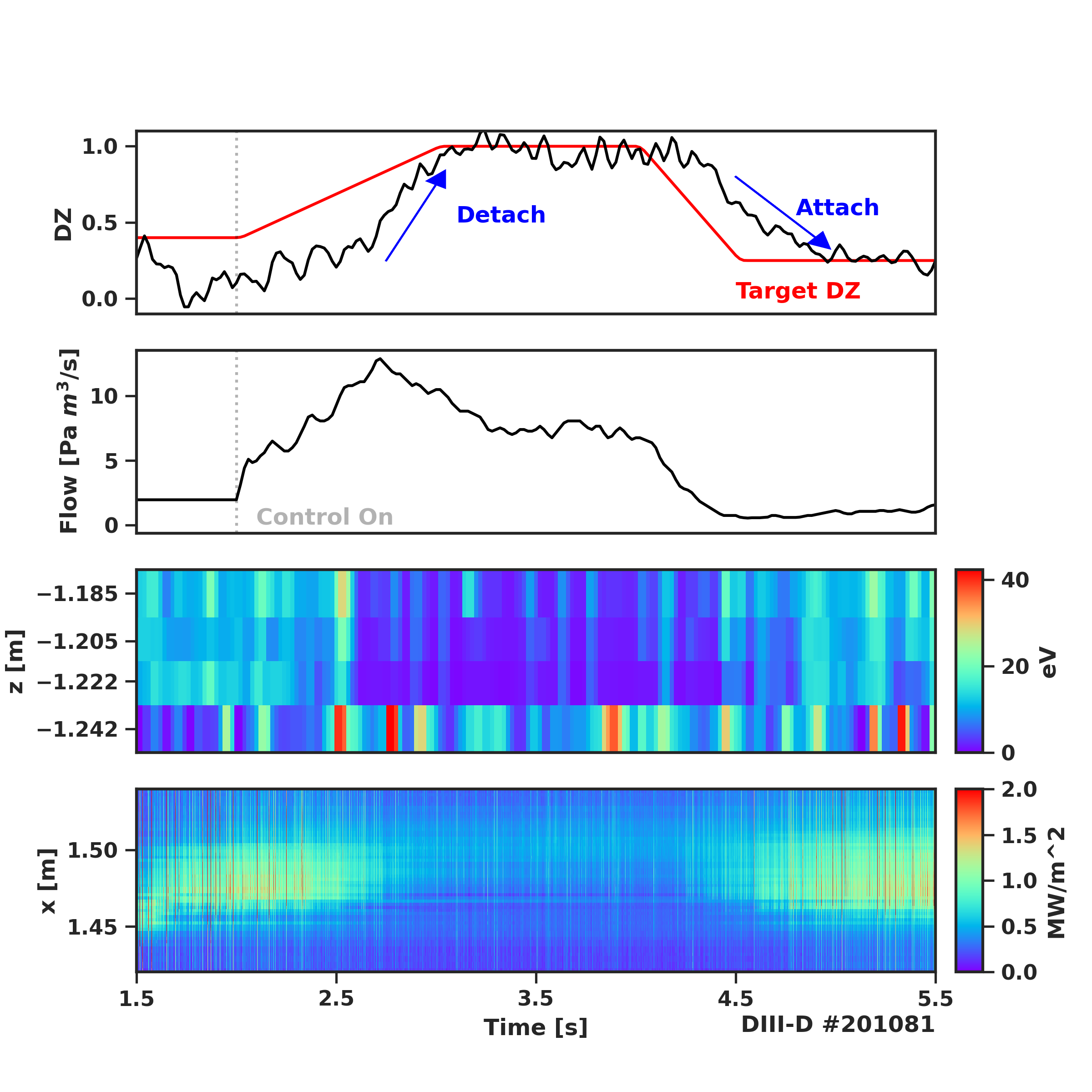}
    \caption{H-mode control of detachment on shot 201081 with $DZ_{Hist}$ using $Z_{E,Hist}$. The decreased heat flux to the divertor surface can be validated with divertor Thomson chords 2,3 which reduces from 20eV to 3eV. Similarly, IRTV shows reduced heat flux to the surface of the divertor with $DZ$ near 1.}
    \label{fig:201081}
\end{figure}

Auxiliary power from NBI was increased to 4MW under the same reference shape and B-field as the L-mode shot, where $DZ$ was increased, then decreased to test control over detachment. In addition, the model used for $Z_E$ prediction was updated to $Z_{E,Hist}$. Figure \ref{fig:201081} shows the resulting $DZ_{Hist}$ and corresponding divertor Thomson scattering (DTS) and Infrared TV (IRTV) measurements \cite{eldon_initial_2012}. Chords at -1.222m and -1.205m measure detachment with electron temperature $T_e$ cliff at around 2650ms and 4450ms. The bifurcation occurs around when $DZ$ is 0.5. It is commonly defined as a rapid evolution of the electron temperature and associated radiation front from near the target upwards to the edge of the separatrix near the X-point \cite{jaervinen_eifmmodetimeselsetexttimesfib_2018}. The mechanism behind this cliff has been posited to be due to the $E\times B$ drift-driven particle sink with ion $B\times \nabla B$ directed into the divertor \cite{rognlien_comparison_2017}. IRTV measurement of surface heating shows a consistent drop in surface energy flux to 0.55MWm$^{-2}$ to the divertor as $DZ$ is high. The mean absolute difference between the target and signal for the raw signal was 5.87\%. After adjusting for lag and starting with the first detachment, was 2.04\%.

\subsection{Active Feedback Control in  H-Mode with Varying Equilibrium}

\begin{figure}
    \includegraphics[width=\columnwidth]{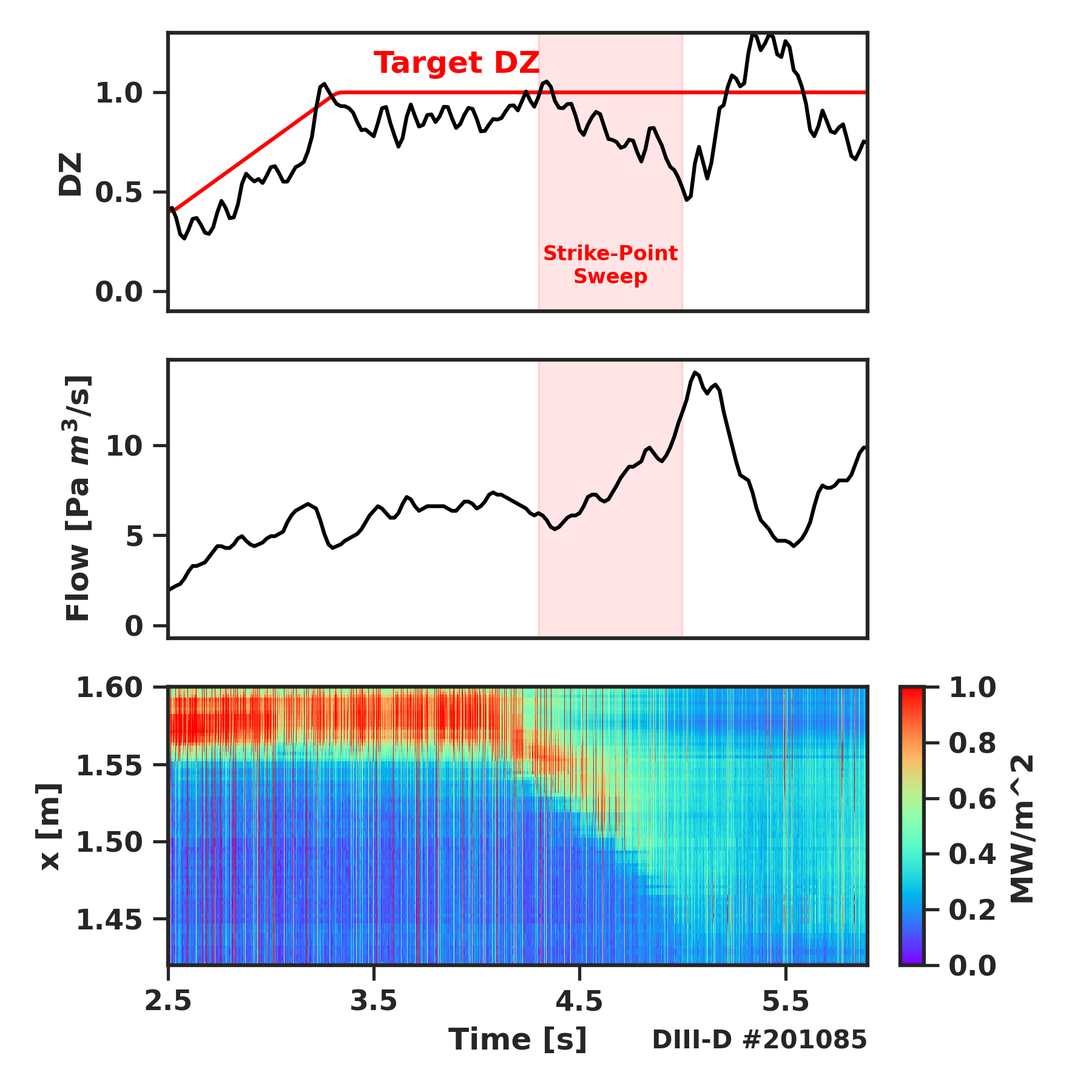}
    \caption{H-mode control of 201085 with $DZ_{Hist}$ demonstrates that the observed $DZ_{Hist}$ signal is influenced by a radially moving emission front while it should ideally remain the same. In addition, IRTV measurements show that initial divertor heat flux remains high even though $DZ$ is near 1. Then radiation decreases due to the increasing gas from a higher $DZ$ prediction as seen with the strike point radial outwards sweep. This causes gas to be puffed in, causing detachment.}
    \label{fig:201085}
\end{figure}

Shot 201085 continues to use $DZ_{Hist}$ for active feedback control. Feedback control is shown in Figure \ref{fig:201085}. This time, after $DZ$ is held steady at 1.0, the X point and corresponding strike point are moved from 1.52m to 1.42m in the poloidal direction. Immediately following this shift, $DZ$ signal decreases from 1.0 to 0.45, which causes gas feedback to increase until $DZ$ target overshoots. As seen in the IRTV profile, the average power deposition on the divertor decreases to 0.7$MWm^{-2}$

\subsection{Post-Shot Analysis and $DZ$ Adjustment}

Shot 201084 showed that $Z_{E,Hist}$ is not enough to provide consistent $DZ$ values between 0 and 1. A brighter image can cause a larger resulting dot product such that $DZ$ goes beyond the intended range.

To maintain consistency across all intensities, the image intensity can be normalized by subtracting the mean intensity and dividing by the standard deviation for each frame. This produces $Z_{E,Norm}$ and its corresponding $DZ_{Norm}$

Shot 201085 shows that even with normalization, a radial repositioning of the X-point will cause $DZ_{Norm}$ to change. This is likely due to the viewing distance of the plasma ring to the camera. As the plasma moves radially outwards, the overlapping pixels with the weight decreases, causing the overall $DZ$ to drop. This effect seems to scale linearly.

\begin{figure}
\centering
\subfigure[]{\includegraphics[width=0.49\textwidth]{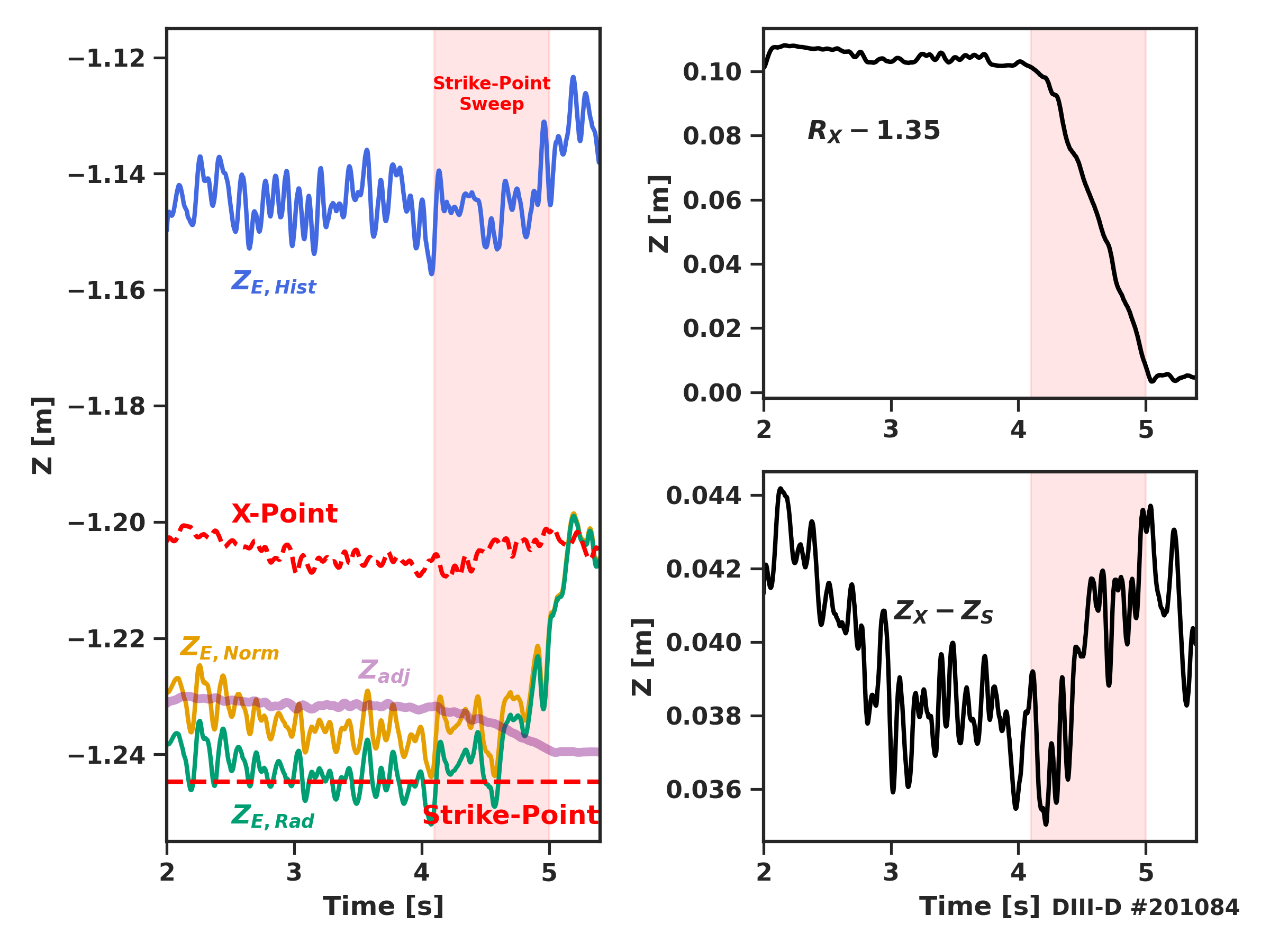}} 
\subfigure[]{\includegraphics[width=0.49\textwidth]{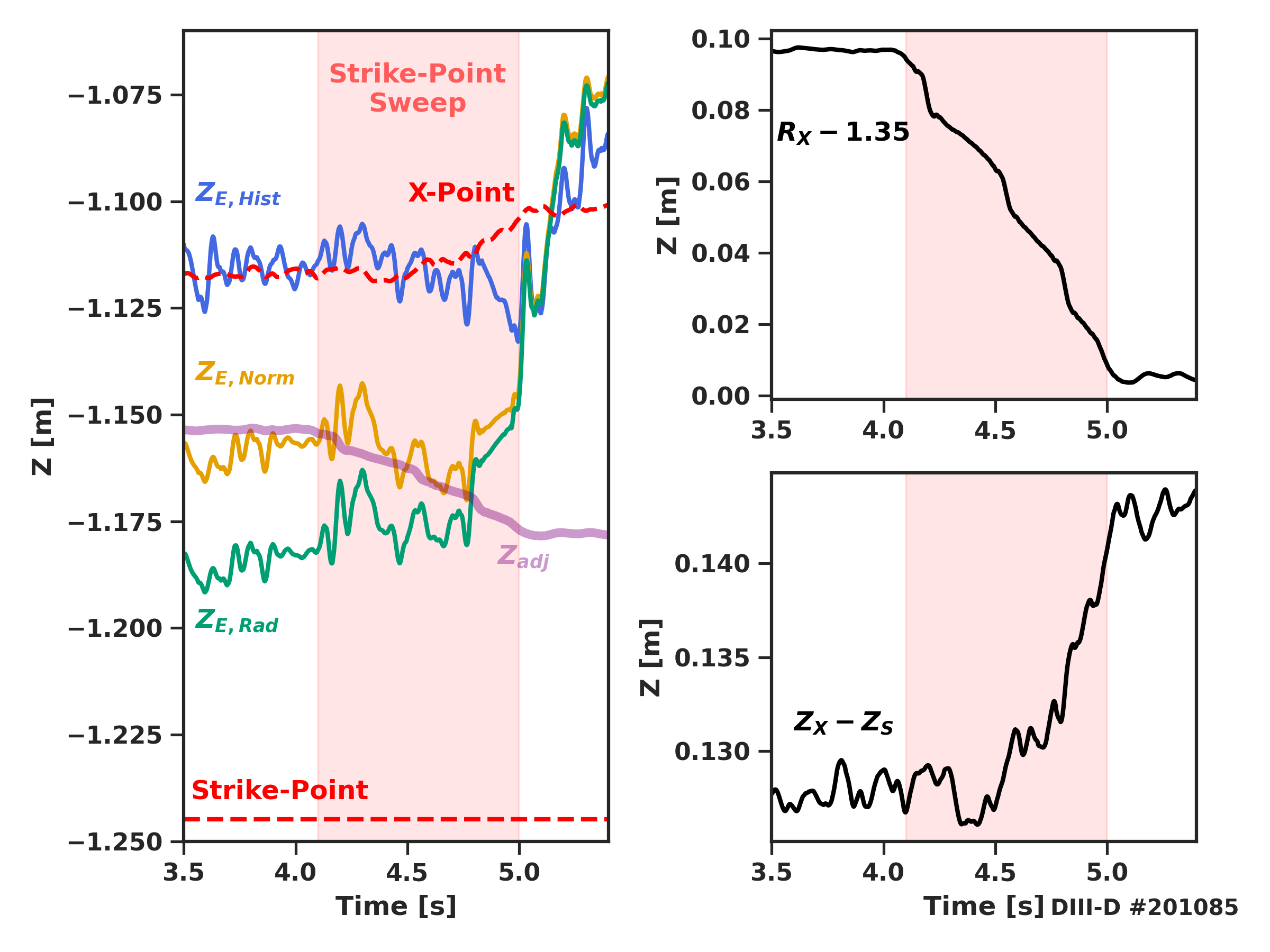}}
\caption{Determining the adjustment to $Z_{E,Norm}$. $Z_E$ should not go below the strike point in steady state, and should not go much higher than the $X$ point. Equation \ref{eq:Zadj} shows this adjustment, with the scaling factor determined by matching the slope of $Z_{adj}$ to the slope of $Z_{E,Norm} $ in (b). This scaling causes $Z_{E,Rad}$ to reach near the strike point, transiently dipping below it in (a) due to signal noise.}
\label{fig:ez_adjust}
\end{figure}

To account for the change due to radial position, the difference between the radial position of the X Point from EFIT and the radial position of the X Point when the strike point is near the edge of the lower divertor shelf is taken. This difference uses the X-point instead of the strike point in order to integrate with the PCS. Then this adjustment value can be subtracted from $Z_{E,Norm}$. However, shot 201084 shows that when $Z_{E,Norm}$ is already near the strike point, the subtraction can cause the adjusted $Z_{E,Norm}$ to go below the strike point which returns negative values of $DZ_{Norm}$. To compensate for this such that the value $Z_E$ minimally adjusts when near the strike point, the adjustment value is multiplied by the difference between $Z_X$ and $Z_S$ so that as the leg becomes shorter, the adjustment factor decreases. Then an empirical fit is then made to the strike-point sweep region for 201085 between $Z_{E,Norm}$ as predicted by the normalized model, and then validated across all shots during the campaign to make sure the new values of $Z_E$ are reasonable, with lower bounds at the strike point, with an example shown in Figure \ref{fig:ez_adjust}.

This finally provides $Z_{E,Rad}$ and its corresponding $DZ_{Rad}$ which keeps $DZ$ consistent with radial changes to the X-Point.

\begin{figure}
\centering
\subfigure[]{\includegraphics[width=0.49\textwidth]{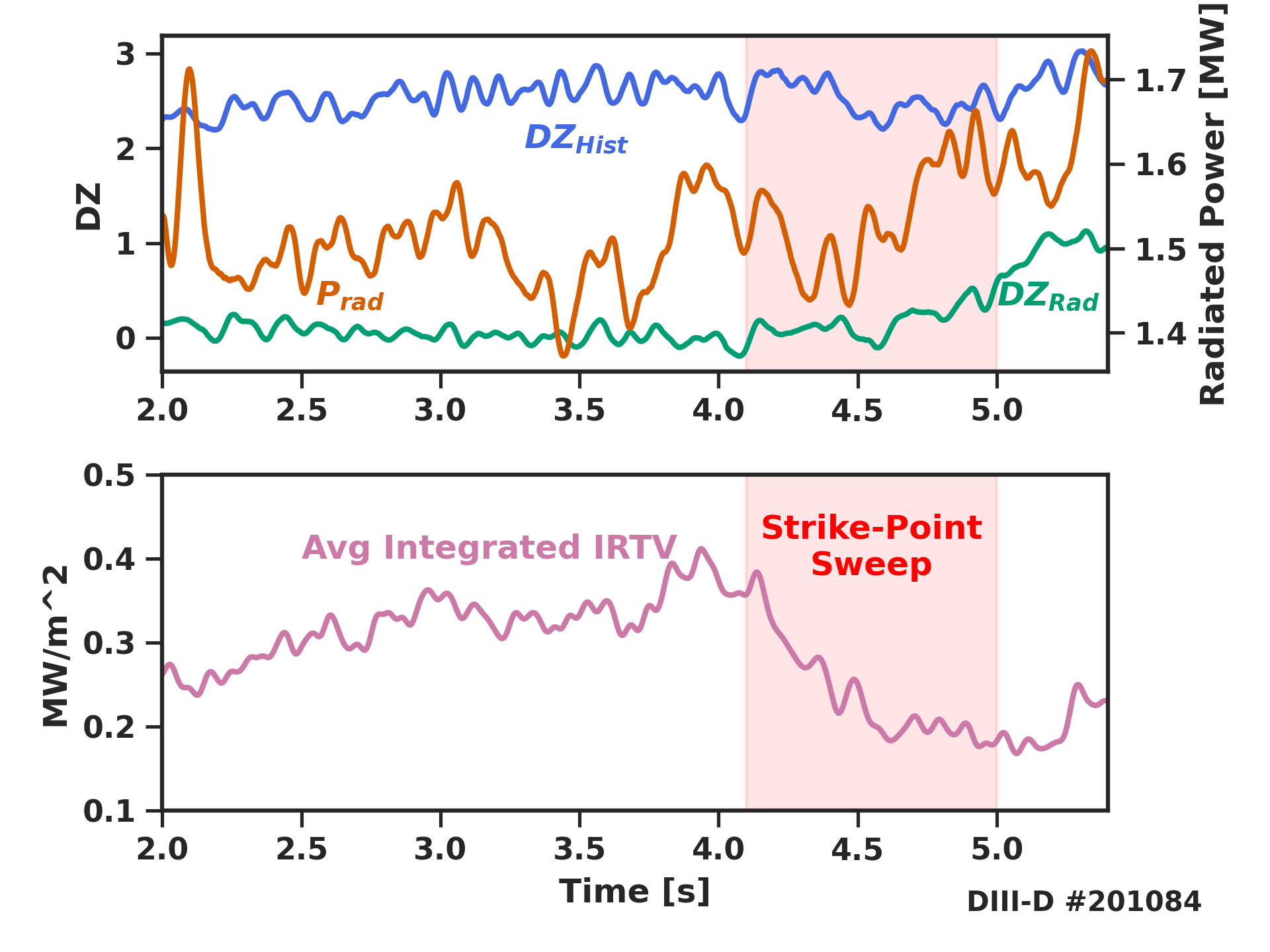}} 
\subfigure[]{\includegraphics[width=0.49\textwidth]{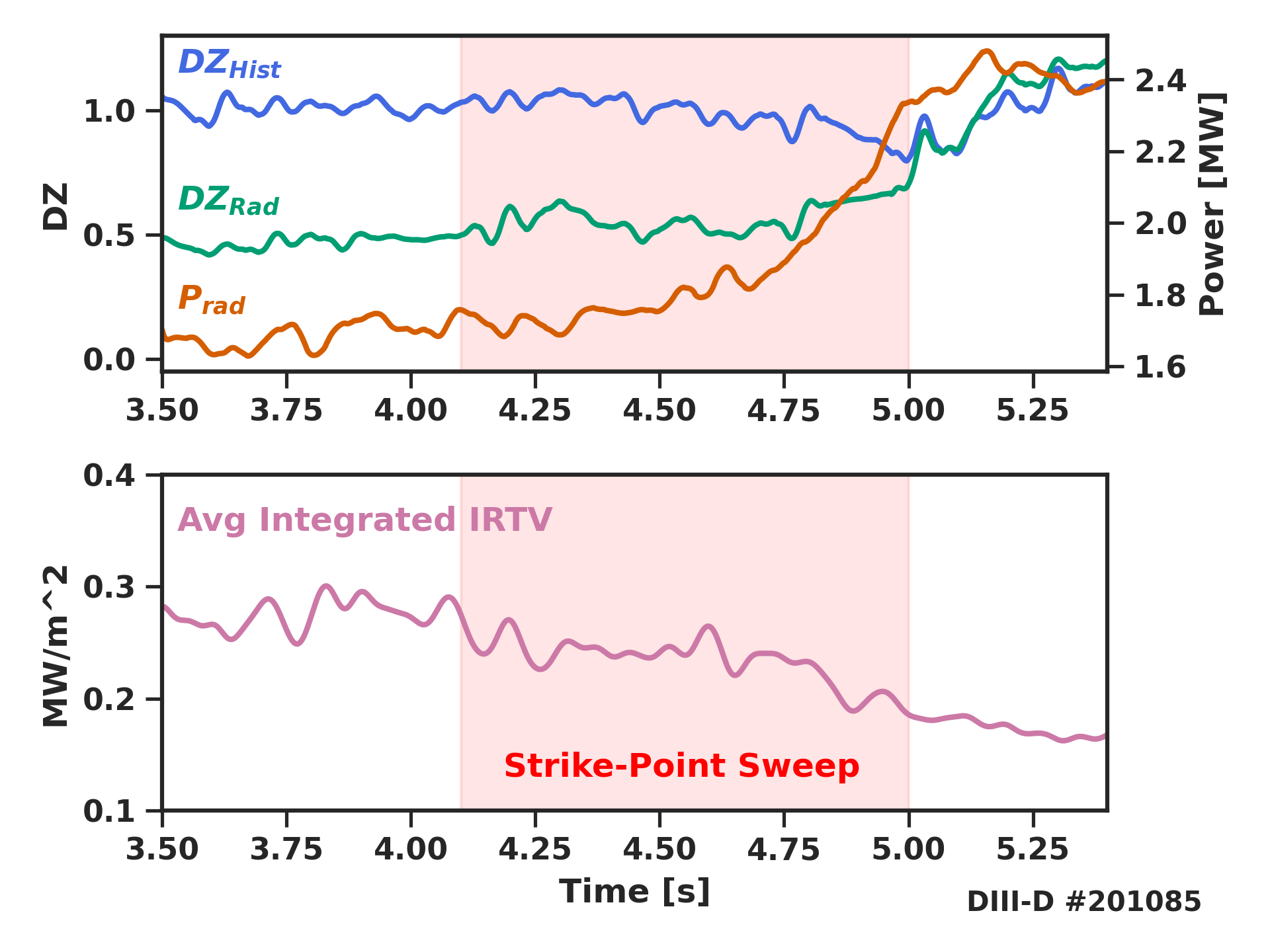}}
\caption{Validation of $DZ_{Rad}$ shows correction for the strike point sweep model inconsistency. $DZ_{Rad}$ and $P_{rad}$ (i.e. $P_{rad,divl}$) scale similarly. Validation with IRTV temperature especially during the strike point sweep shows that as $DZ$ increases to that near the X point, the divertor temperature drops.}
\label{fig:dz_adjust}
\end{figure}

The new $DZ_{Rad}$ is given by

\begin{equation}
    DZ_{Rad} = 1 - \frac{Z_X - (Z_{E,Norm} - Z_{adj})}{Z_X - Z_S}
    \label{eq:rdz}
\end{equation}

\begin{equation}
    Z_{adj} = \alpha*(R_X - R_{edge}) * (Z_X - Z_S)
    \label{eq:Zadj}
\end{equation}

where $R_{edge} = 1.35m$ and the empirical adjustment factor $\alpha = 2.1$, shown in Figure \ref{fig:dz_adjust}. A summary of the $DZ$ models and when they were developed in relation to the experiments can be found in Figure \ref{fig:preprocessing} and \ref{tab:t2}.

\begin{table*}[t]
\centering
\begin{tabular}{|l|l|}
\hline
Name         & Modification                                          \\ \hline
$Z_{E,Base}$ & None                                                                      \\ \hline
$Z_{E,Hist}$ & + Fit train hist to avg RT hist of shots 200963-200978      \\ \hline
$Z_{E,Norm}$ & + Train \& RT intensity $\mu = 0$ and $\sigma = 1$                                  \\ \hline
$Z_{E,Rad}$  & + Train \& RT $Z_E$ minus $Z_{adj}$, in Eq \ref{eq:Zadj} \\ \hline
\end{tabular}
\caption{Detailing differences in $Z_E$}
\label{tab:t2}
\end{table*}

\subsection{Metric Consistency and Benchmarks}

\begin{figure*}[t]
\centering
\subfigure[]{\includegraphics[width=\columnwidth]{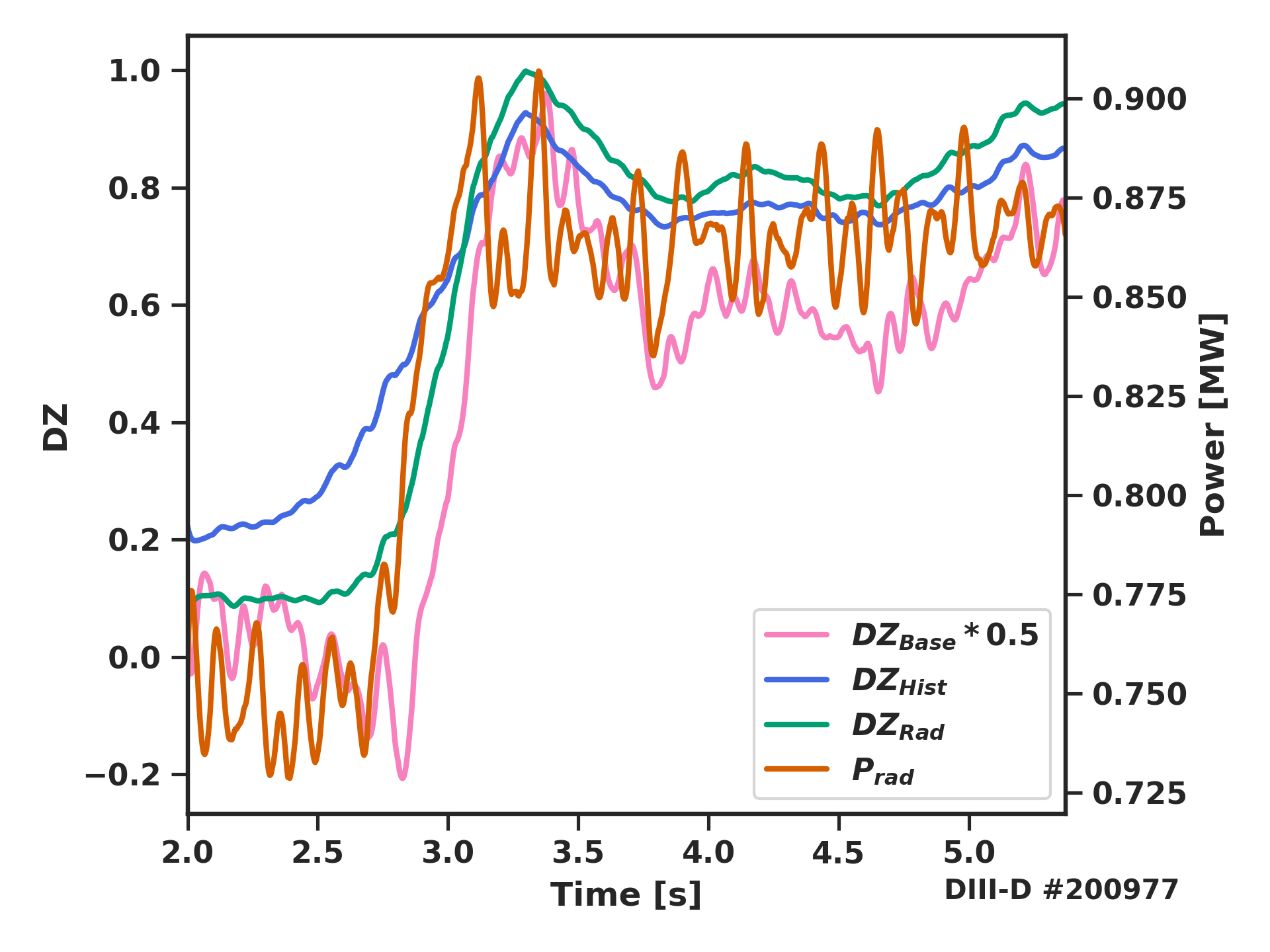}}
\subfigure[]{\includegraphics[width=\columnwidth]{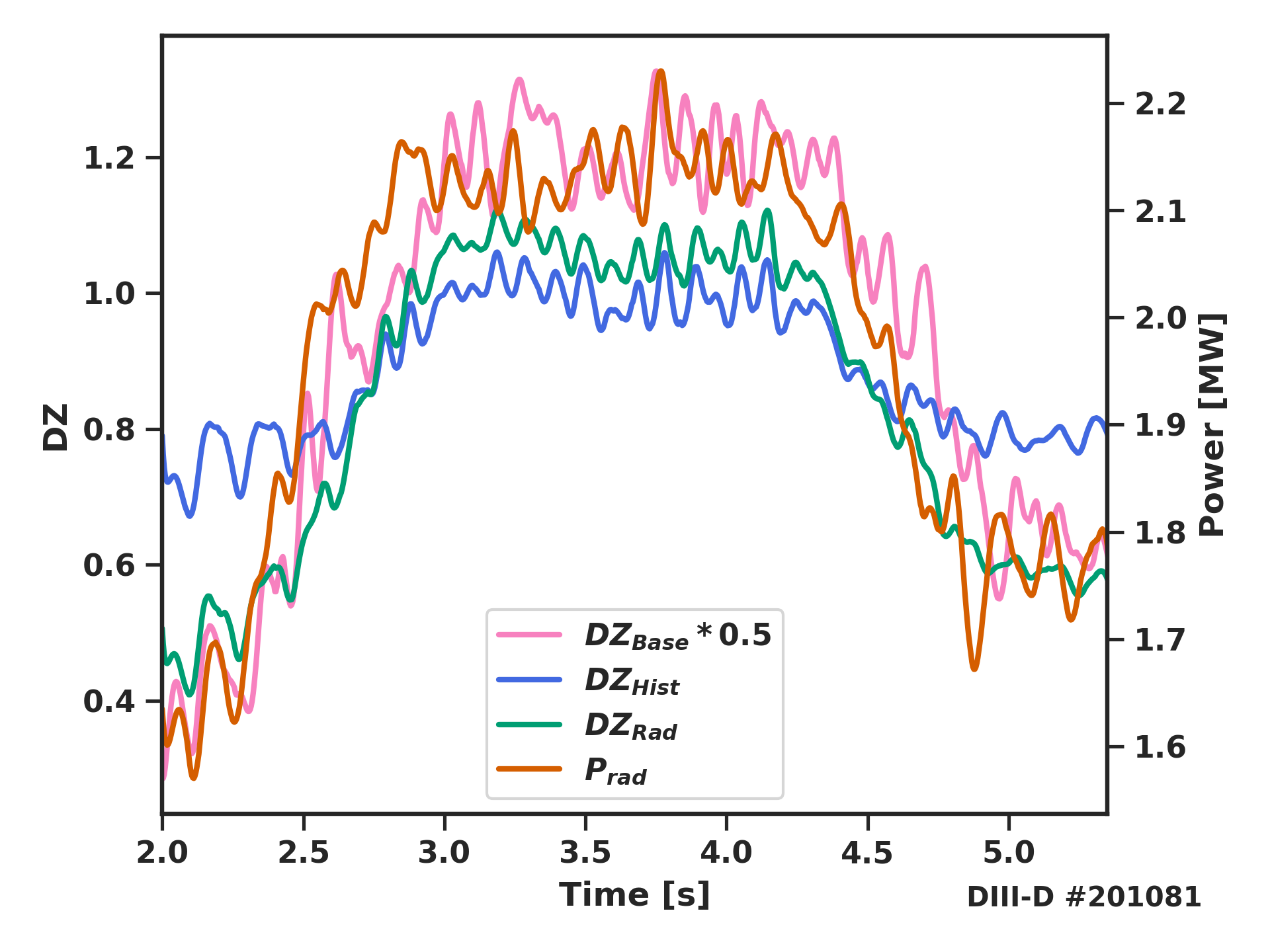}}
\subfigure[]{\includegraphics[width=\columnwidth]{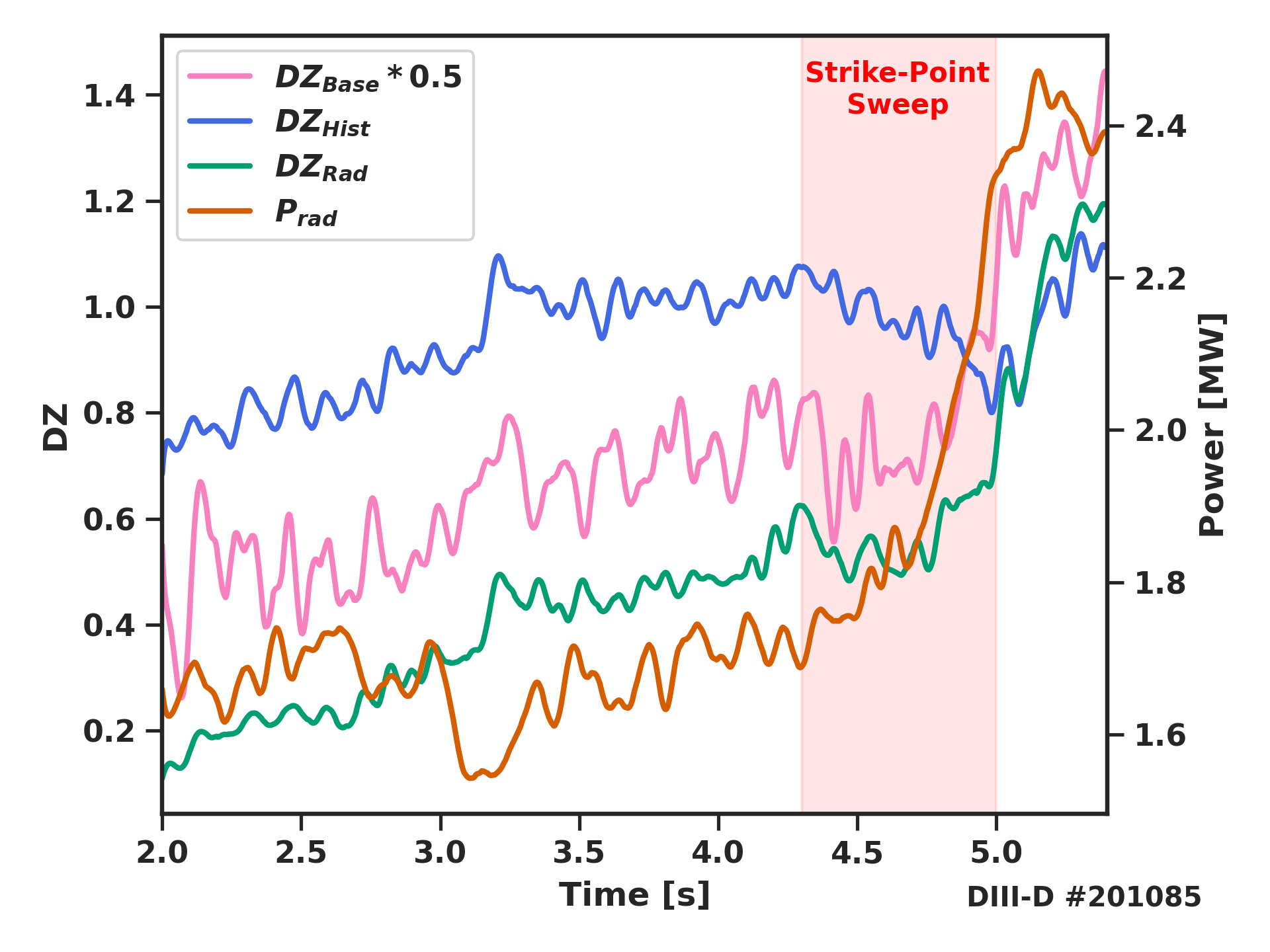}}
\subfigure[]{\includegraphics[width=\columnwidth]{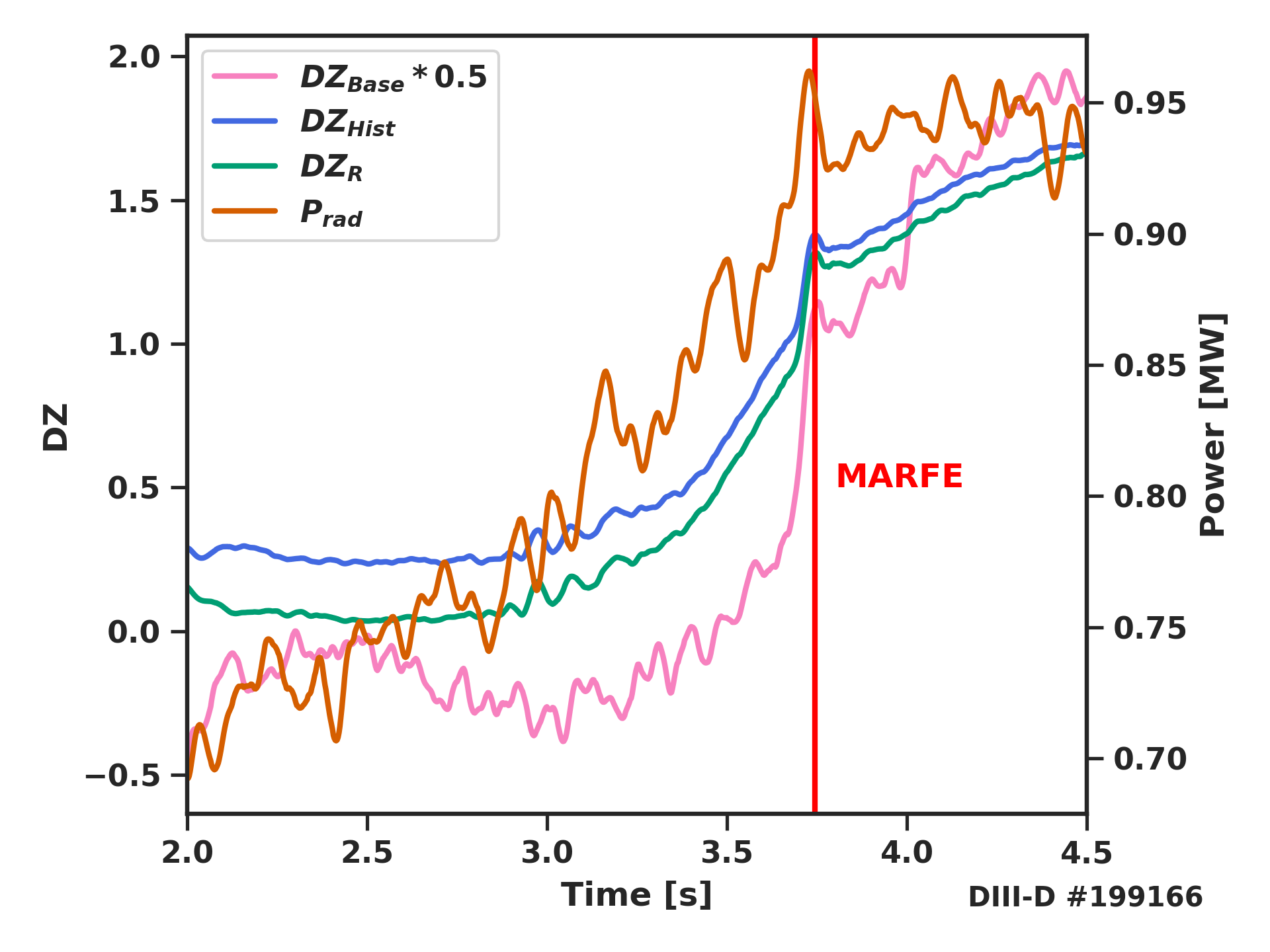}}
\caption{Comparing $DZ$ models with $P_{rad}$ (i.e. $P_{rad,divl}$), the heat flux radiated in the lower divertor, show similar predictions trends (a) DIII-D 200977 L-Mode: $DZ_{Rad}$ has less noise compared to $P_{rad}$ (b) DIII-D 201081 H-Mode: $DZ_{Rad}$ has a similar noise profile to $P_{rad}$ (c) DIII-D 201085 H-Mode: $DZ_{Rad}$ corrects for the decrease in $DZ_{Hist}$ during strike point sweep, which is similar to the trend in $P_{rad}$.(d) DIII-D 199166 L-Mode with observation of MARFE at 3.705 seconds.}
\label{fig:mdlcompare}
\end{figure*}

Comparisons between the models show that matching histograms can produce stronger signals, and that radial scaling adjusts for radial X point movement. Figure \ref{fig:mdlcompare} shows comparisons between the different models.

\begin{figure*}[t]
\centering
\subfigure[]{\includegraphics[width=\columnwidth]{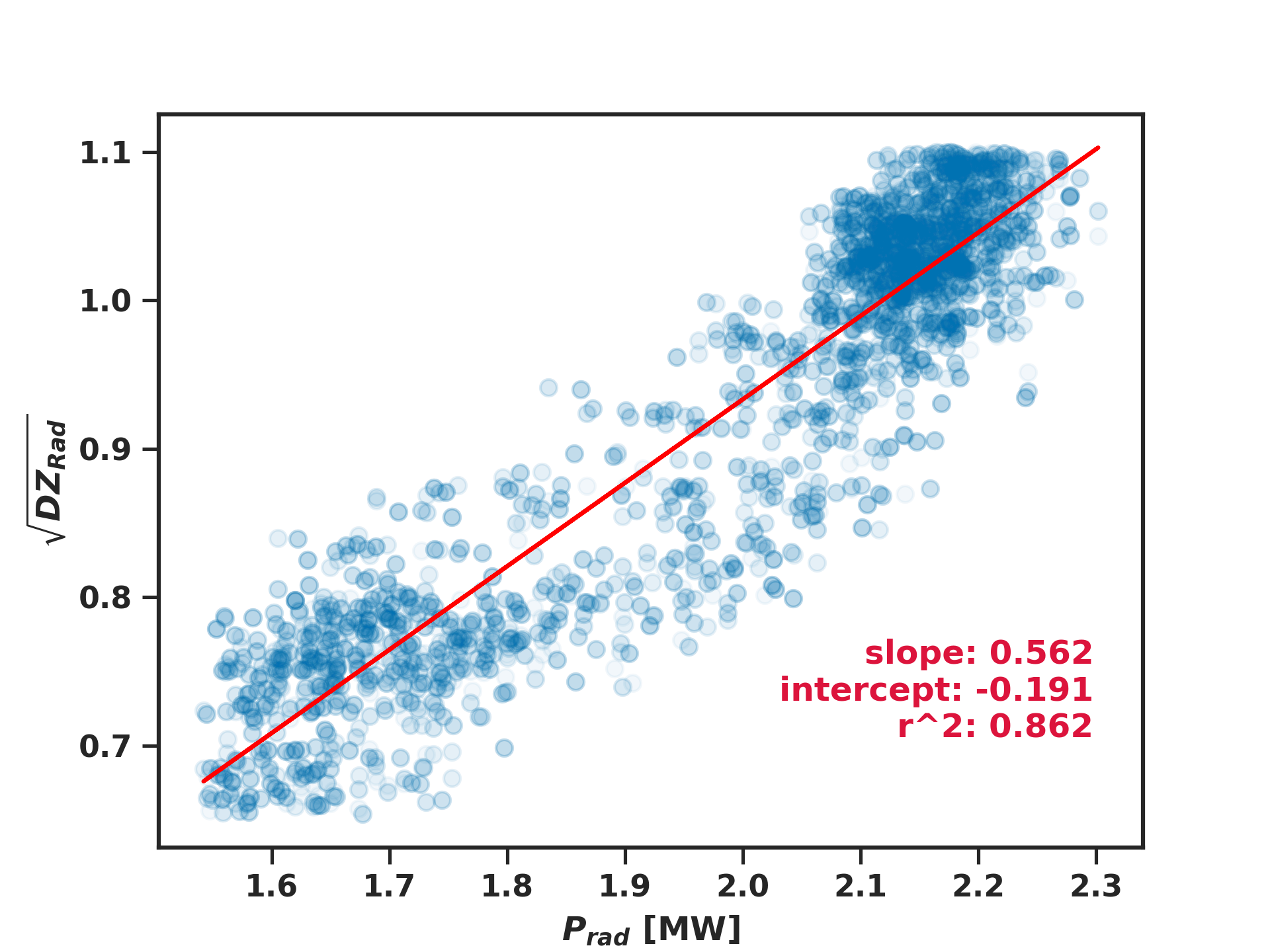}}
\subfigure[]{\includegraphics[width=\columnwidth]{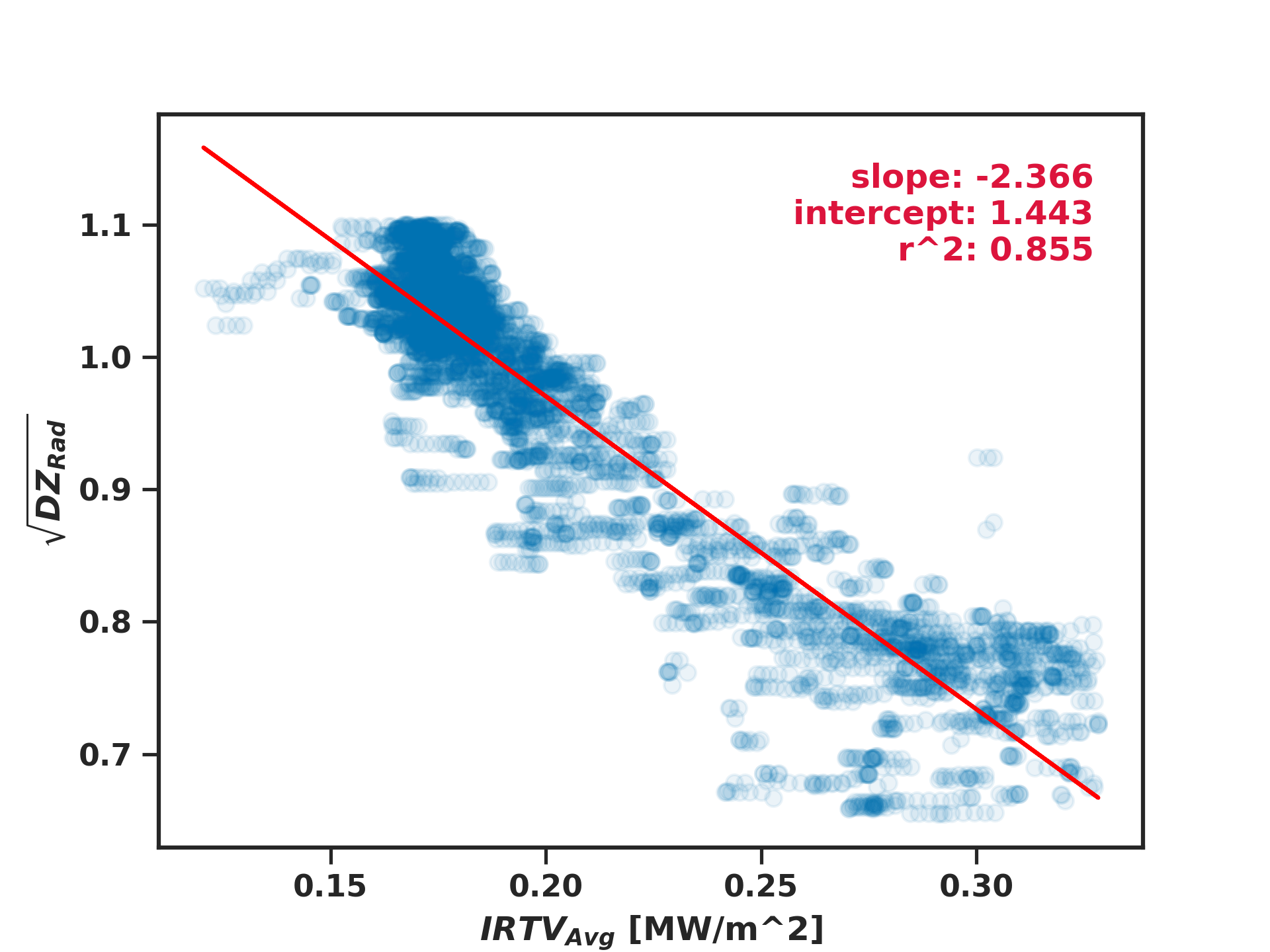}}
\caption{Correlating the square root of $DZ_{Rad}$ with (a) $P_{rad}$ (b) Average $IRTV$. Outliers of $Z > 2$ were removed, and the window was set to $0.65 < \sqrt{DZ_{Rad}} < 1.1$ which corresponds to the area above where the plasma detaches and before it goes above the X point. This analysis combines the results of DIII-D \# 201074 - 201083.}
\label{fig:dzcorrelate}
\end{figure*}

As a metric, $DZ_{Rad}$ provides a consistent value that resembles a normalized $P_{rad}$ for both L-mode and H-mode shots. Since all experiments are conducted with lower divertor detachment, $P_{rad}$ refers to $P_{rad,divl}$, or total radiated power in the lower divertor region. $P_{rad}$ increases earlier than $DZ_{Rad}$ with a maximum delay of roughly 50 ms in transient states before reaching a similar peak. It then decreases later than $DZ_{Rad}$, leading with an approximately 50 ms head start. The profile distributions indicate that this discrepancy is likely due to C-III emission front's source at the divertor plate. Further correlation shows that this may be more likely due to $DZ$ having a squared effect of $P_{rad}$, as when $DZ_{Rad}$ is squared for regions corresponding to a non-detached state, it is linearly correlated to $P_{rad}$ and the heat flux of the $IRTV$, shown in Figure \ref{fig:dzcorrelate}. This is intended to provide more consistent fine-control over mid-range $DZ$ values.

\subsection{Limitations}

The discrepancy in radiation front position corresponding to detachment may be due to gas penetration timescales. Gas puffing has an approximate delay of approximately 200 ms before becoming effective. Additional delays from image interlacing and data processing can result in a noticeable total control delay. Future work can employ higher baseline gas commands for faster response time.

To explain the delay in upwards movement and faster downward movement when comparing $DZ$ to $P_{rad}$, $P_{rad}$ integrates all the radiating species and charge states in the divertor, while TangTV is only looking at one species (C) and one charge state (2+).

\section{Summary and Outlook}

We have demonstrated a successful real-time camera control system at DIII-D that utilizes an AI-based linear controller to precisely control divertor detachment. A regression model provided accurate predictions and effective control over the full lower divertor region with a single diagnostic, offering a promising pathway for implementation in ITER and other future fusion devices with tangential divertor viewing systems. This approach relies on detecting an emission that accurately reflects the radiation front location, a condition that is readily met for carbon-based systems but becomes more challenging for devices using heavier atoms like Ne or Ar, whose emissions predominantly occur in the UV. Notably, detachment characterization using the $D_2$ Fulcher band at MAST-U suggests that visible detachment control is a viable method that could extend to most fusion plasmas \cite{wijkamp_characterisation_2023}.

Future work should investigate regions near $DZ = 0.5$, building on previous partial detachment measurements. Incorporating additional diagnostics, such as bolometric readings, into the ML algorithm may further enhance prediction accuracy. Faster integration with the plasma control system (PCS), potentially via GPU acceleration, is essential for real-time applications, and expanding the training dataset will improve model robustness. Testing the control system under higher confinement regimes, such as ITER baseline scenarios with increased ELM activity, is a logical next step; in these scenarios, integrating ELM mitigation or suppression techniques will be critical, especially given earlier work suggesting that camera systems can predict disruptions.

Our results confirm that detachment control can be achieved with a single camera system and simple AI-based linear modeling. This work highlights the potential of classical linear control techniques, trusted for their stability bounds and interpretability in managing complex, nonlinear phenomena \cite{otte_safe_2013}. While advanced neural networks may be necessary for certain tasks, a thorough analysis using classical tools should always be considered first. Additionally, the successful use of average C-III emission height as a proxy for the emission front opens up avenues for applying ML-enabled real-time control to other 2D imaging diagnostics, such as imaging bolometers or shape control systems.

\ack
The authors thank all the DIII-D team members for their support. This work is supported by US DOE Grant Nos. DE-FC02-04ER54698 and DE-SC0024527. This work is supported by LLNL, DE-AC52-07NA27344. The authors also gratefully acknowledge financial support from the Princeton Laboratory for Artificial Intelligence under Award 2025-97.

Disclaimer: This report was prepared as an account of work sponsored by an agency of the United States Government. Neither the United States Government nor any agency thereof, nor any of their employees, makes any warranty, express or implied, or assumes any legal liability or responsibility for the accuracy, completeness, or usefulness of any information, apparatus, product, or process disclosed, or represents that its use would not infringe privately owned rights. Reference herein to any specific commercial product, process, or service by trade name, trademark, manufacturer, or otherwise does not necessarily constitute or imply its endorsement, recommendation, or favoring by the United States Government or any agency thereof. The views and opinions of authors expressed herein do not necessarily state or reflect those of the United States Government or any agency thereof.

Special thanks to R. Reed and R. Shousha for making this all possible.

\section*{References}

\bibliographystyle{unsrt}
\bibliography{references}

\end{document}